\documentclass[balance=false,sigconf,authorversion]{acmart}

\copyrightyear{2021}
\acmYear{2021}
\setcopyright{acmlicensed}
\acmConference[KDD '21]{Proceedings of the 27th ACM SIGKDD Conference on Knowledge Discovery and Data Mining}{August 14--18, 2021}{Virtual Event, Singapore}
\acmBooktitle{Proceedings of the 27th ACM SIGKDD Conference on Knowledge Discovery and Data Mining (KDD '21), August 14--18, 2021, Virtual Event, Singapore}
\acmPrice{15.00}
\acmISBN{978-1-4503-8332-5/21/08}
\acmDOI{10.1145/3447548.3467231}

\usepackage[ruled]{algorithm2e}
\usepackage{float}
\usepackage{hyperref}
\usepackage{mathtools}
\usepackage{subcaption}

\newcommand{\rocket}{\textsc{Rocket}}
\newcommand{\minirocket}{\textsc{MiniRocket}}

\begin{document}

\fancyhead{}

\title{{\minirocket}}
\subtitle{A Very Fast (Almost) Deterministic Transform for Time Series Classification}

\author{Angus Dempster}
\email{angus.dempster1@monash.edu}
\affiliation{
  \institution{Monash University}
  \city{Melbourne}
  \country{Australia}
}

\author{Daniel F. Schmidt}
\email{daniel.schmidt@monash.edu}
\affiliation{
  \institution{Monash University}
  \city{Melbourne}
  \country{Australia}
}

\author{Geoffrey I. Webb}
\email{geoff.webb@monash.edu}
\affiliation{
  \institution{Monash University}
  \city{Melbourne}
  \country{Australia}
}

\begin{abstract}
  {\rocket} achieves state-of-the-art accuracy for time series classification with a fraction of the computational expense of most existing methods by transforming input time series using random convolutional kernels, and using the transformed features to train a linear classifier.  We reformulate {\rocket} into a new method, {\minirocket}.  {\minirocket} is up to $75$ times faster than {\rocket} on larger datasets, and almost deterministic (and optionally, fully deterministic), while maintaining essentially the same accuracy.  Using this method, it is possible to train and test a classifier on all of 109 datasets from the UCR archive to state-of-the-art accuracy in under 10 minutes.  {\minirocket} is significantly faster than any other method of comparable accuracy (including {\rocket}), and significantly more accurate than any other method of remotely similar computational expense.
\end{abstract}



\maketitle

\section{Introduction} \label{sec-introduction}

\begin{figure}
\centering
\includegraphics[width=\linewidth]{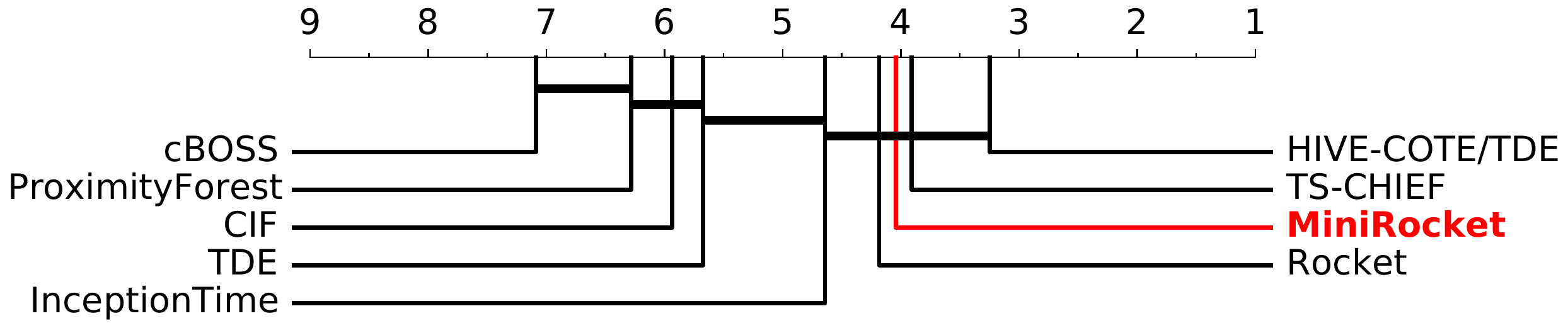}
\caption{Mean rank of {\minirocket} in terms of accuracy versus other SOTA methods over 30 resamples of 109 datasets from the UCR archive.}
\Description[In terms of accuracy, Apricot ranks just ahead of Rocket, but behind both TS-CHIEF and HIVE-COTE/TDE]{In terms of accuracy, Apricot ranks just ahead of Rocket, but behind both TS-CHIEF and HIVE-COTE/TDE.  Apricot is in the same clique as InceptionTime, Rocket, TS-CHIEF, and HIVE-COTE, that is, the pairwise differences between these classifiers are not statistically significant.}
\label{fig-rank-ucr109}
\end{figure}

\begin{figure}
\centering
\includegraphics[width=\linewidth]{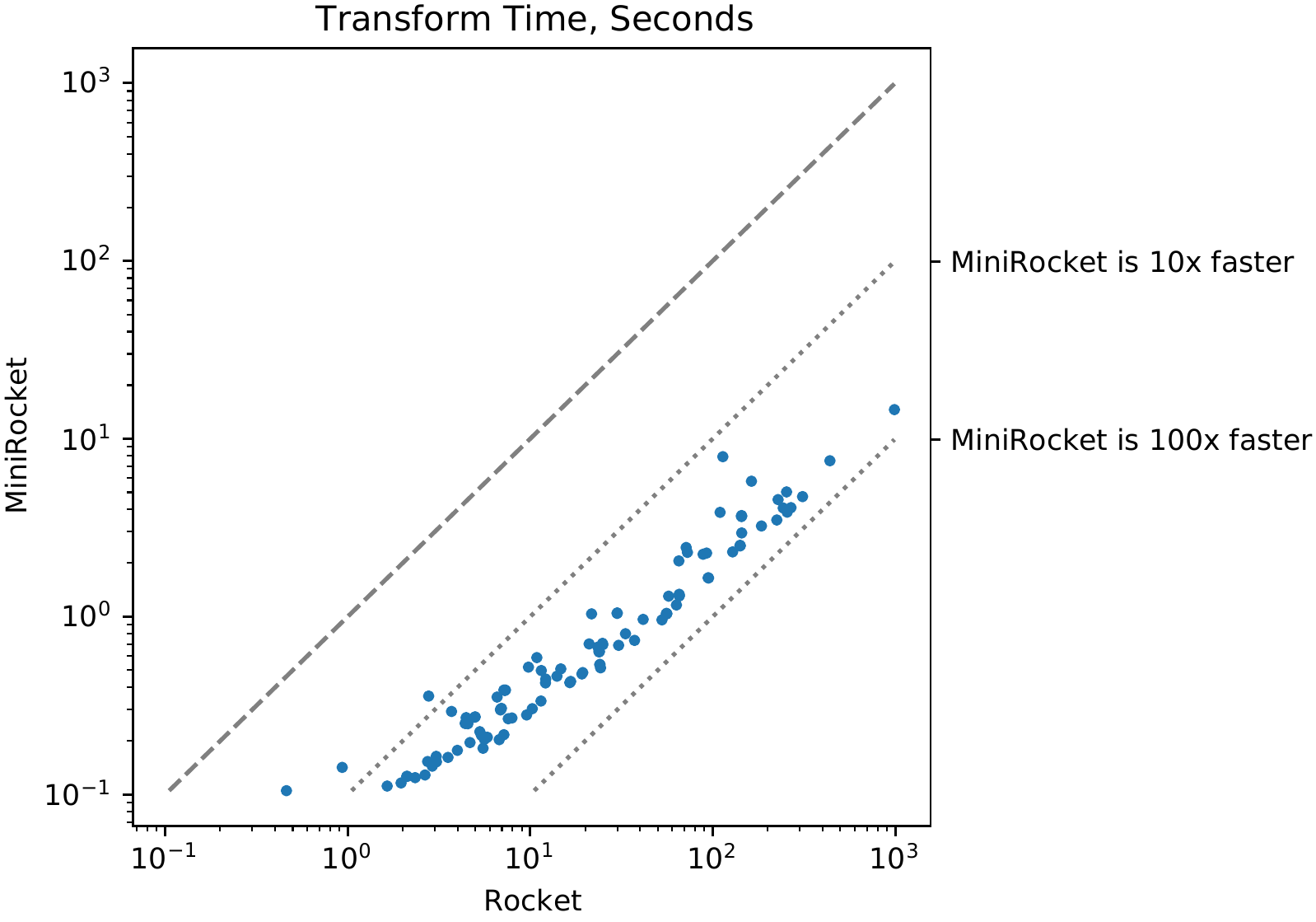}
\caption{Transform time for {\minirocket} versus {\rocket} for the same 109 datasets from the UCR archive.}
\Description[In terms of transform time, Apricot is between approximately 5 and 70 times faster than Rocket]{Scatter plot showing the total transform time (training and test) for Apricot against the total transform time for Rocket for 109 datasets from the UCR archive, on a log scale.  In terms of total transform time, Apricot is between approximately 5 and 70 times faster than Rocket.  The difference in total transform time increases, that is, Apricot is relatively faster, as total transform time increases.  For shorter transform times, Apricot is closer to 10 times faster than Rocket, for longer transform times, Apricot is approaching 70 times faster than Rocket.}
\label{fig-transform-time-minirocket-vs-rocket}
\end{figure}

Until recently, the most accurate methods for time series classification were limited by high computational complexity.  While there have been considerable advances in recent years, computational complexity and a lack of scalability remain persistent problems.

{\rocket} \citep{dempster_etal_2020} achieves state-of-the-art accuracy with a fraction of the computational expense of any method of comparable accuracy by transforming input time series using random convolutional kernels, and using the transformed features to train a linear classifier.  We show that it is possible to reformulate {\rocket}, making it up to $75$ times faster on larger datasets, and making it almost entirely deterministic (and optionally, with additional computational expense, fully deterministic), while maintaining essentially the same accuracy.  We call this method {\minirocket} (for \textbf{MINI}mally \textbf{R}and\textbf{O}m \textbf{C}onvolutional \textbf{KE}rnel \textbf{T}ransform).

Like {\rocket}, {\minirocket} transforms input time series using convolutional kernels, and uses the transformed features to train a linear classifier.  However, unlike {\rocket}, {\minirocket} uses a small, fixed set of kernels, and is almost entirely deterministic.  {\minirocket} maintains the two most important aspects of {\rocket}: dilation and PPV, i.e., `proportion of positive values' pooling \citep{dempster_etal_2020}.  {\minirocket} exploits various properties of the kernels, and of PPV, in order to massively reduce the time required for the transform.  {\minirocket} demonstrates that, while random convolutional kernels are highly effective, it is possible to achieve essentially the same accuracy using a mostly-deterministic and much faster procedure.

Figure \ref{fig-rank-ucr109} shows the mean rank of {\minirocket} in terms of accuracy versus other state-of-the-art methods over 30 resamples of 109 datasets from the UCR archive of benchmark time series \citep{dau_etal_2019}.  On average, {\minirocket} is marginally more accurate than {\rocket}, and slightly less accurate than the most accurate current methods.

\begin{sloppypar}
  Figure \ref{fig-transform-time-minirocket-vs-rocket} shows total transform time (training and test) for {\minirocket} versus {\rocket}, for the same 109 datasets.  On average, {\minirocket} is more than $30$ times faster than {\rocket} (the advantage is even greater for larger datasets: see Section \ref{subsec-scalability}).  Restricted to a single CPU core, total compute time for {\rocket} over all 109 datasets is 2 hours 2 minutes (1h 55m transform time), versus just 8 minutes for {\minirocket} (2m 30s transform time).  To put this in context, total compute time for {\minirocket} for all 109 datasets is less than the compute time for {\rocket} for just one of those datasets.  (Compute times are averages over 30 resamples, run on a cluster using Intel Xeon E5-2680 v3/4 and Xeon Gold 6150 CPUs, restricted to a single CPU core per dataset per resample.)
\end{sloppypar}

While only broadly comparable due to hardware and software differences, total compute time for the same 109 datasets using a single CPU thread is approximately 13 hours for cBOSS, more than a day for CIF, more than two days for TDE, approximately a week for Proximity Forest, more than two weeks for HIVE-COTE, and several weeks for TS-CHIEF \citep{bagnall_etal_2020,middlehurst_etal_2020a,middlehurst_etal_2020b}.  Total compute time for InceptionTime (using GPUs, and for the original training/test splits rather than resamples) is more than 4 days \citep{ismailfawaz_etal_2020}.

{\minirocket} represents a significant advance in accuracy relative to computational cost.  {\minirocket} is significantly faster than any other method of comparable accuracy (including {\rocket}), and significantly more accurate than any other method of even roughly-similar computational expense.

The rest of this paper is structured as follows.  In Section \ref{sec-related-work}, we review relevant related work.  In Section \ref{sec-method}, we detail the changes from {\rocket} to {\minirocket}.  In Section \ref{sec-experiments}, we present experimental results for {\minirocket} in terms of accuracy and scalability, as well as a sensitivity analysis in relation to key parameter choices.

\section{Related Work} \label{sec-related-work}

\subsection{Current State of the Art}

Recent advances in accuracy have largely superseded the most accurate methods originally identified in \citep{bagnall_etal_2017}.  According to \citep{bagnall_etal_2020,middlehurst_etal_2020a,middlehurst_etal_2020b}, the most accurate current methods for time series classification are HIVE-COTE and its variants \citep{lines_etal_2018}, TS-CHIEF \citep{shifaz_etal_2020}, InceptionTime \citep{ismailfawaz_etal_2020}, and {\rocket} \citep{dempster_etal_2020}.  However, while accuracy has improved, with some exceptions computational complexity and a lack of scalability remain persistent problems.

TS-CHIEF builds on Proximity Forest, an ensemble of decision trees using distance measures as splitting criteria \citep{lucas_etal_2019}.  In addition to distance measures, TS-CHIEF uses interval-based and spectral-based splitting criteria \citep{shifaz_etal_2020}.

InceptionTime is an ensemble of convolutional neural networks based on the Inception architecture, and is the most accurate convolutional neural network model for time series classification \citep{ismailfawaz_etal_2020}.

The Temporal Dictionary Ensemble (TDE) is a recent dictionary method based on the frequency of occurrence of patterns in time series \citep{middlehurst_etal_2020a}.  TDE combines aspects of earlier dictionary methods including cBOSS \citep{middlehurst_etal_2019}, a more scalable variant of BOSS \citep{schafer_2015}.

Catch22 is a transform based on 22 predefined time series features, used in combination with a decision tree or random forest \citep{lubba_etal_2019}.  On its own, catch22 is fast, but highly inaccurate: see \citep{dempster_etal_2020,middlehurst_etal_2020b}.  The Canonical Interval Forest (CIF) is a recent method which adapts the Time Series Forest (TSF) to use catch22 features \citep{middlehurst_etal_2020b}.  CIF is significantly more accurate than either catch22 or TSF.

\begin{sloppypar}
  HIVE-COTE is an ensemble of other methods including BOSS and TSF.  Two recent variants of HIVE-COTE, namely HIVE-COTE/TDE (using TDE in place of BOSS) and HIVE-COTE/CIF (using CIF in place of TSF) have been shown to be significantly more accurate than HIVE-COTE, or any other existing method for time series classification \citep{middlehurst_etal_2020a,middlehurst_etal_2020b}.  These variants are, in turn, based on an updated `base' version of HIVE-COTE \citep{bagnall_etal_2020}.
\end{sloppypar}

While state of the art in terms of accuracy, with the exception of cBOSS these methods are limited by high computational complexity, requiring days or even weeks to train on the datasets in the UCR archive.  While more scalable, cBOSS is significantly less accurate than most of the other methods.

\subsection{{\rocket}}

{\rocket} achieves state-of-the-art accuracy, matching the most accurate methods for time series classification (with the exception of the most recent variants of HIVE-COTE), but is considerably faster and more scalable than other methods of comparable accuracy \citep{dempster_etal_2020}.

{\rocket} transforms input time series using random convolutional kernels, and uses the transformed features to train a linear classifier.  Each input time series is convolved with $10{,}000$ random convolutional kernels.  {\rocket} applies global max pooling and PPV (for `proportion of positive values') pooling to the resulting convolution output to produce two features per kernel per input time series, for a total of $20{,}000$ features per input time series.  The transformed features are then used to train a linear classifier: a ridge regression classifier, or logistic regression trained using stochastic gradient descent (for larger datasets).

The kernels are random in terms of their length, weights, bias, dilation, and padding: see Section \ref{subsec-removing-randomness}.  The two most important aspects of {\rocket} in terms of achieving state-of-the-art accuracy are the use of dilation, sampled on an exponential scale, and the use of PPV.  {\rocket} forms the basis for {\minirocket}.  The differences between {\rocket} and {\minirocket} are detailed in Section \ref{sec-method}.

\subsection{Other Methods}

The use of a small, fixed set of kernels differentiates {\minirocket} from both {\rocket}, which uses random kernels, and convolutional neural networks such as InceptionTime, which use learned kernels.  It also differentiates {\minirocket} from other methods with at least superficial similarities to {\rocket}, such as random shapelet methods as in \citep{karlsson_etal_2016}, and other random methods such as those based on \citep{rahimi_and_recht_2008}.

In using kernels with weights constrained to two values (see Section \ref{subsubsec-weights}), there are obvious similarities with binary and quantised convolutional neural networks \citep{rastegari_etal_2016,hubara_etal_2018}.  {\minirocket} makes use of at least two advantages of binary/quantised kernels, namely, the ability to perform the convolution operation via addition, as well as efficiencies arising from the relatively small number of possible binary kernels of a given size, e.g., \citep{rastegari_etal_2016,juefeixu_etal_2017,hubara_etal_2018}.  However, while the kernels used in {\minirocket} are binary in the sense of having only two values, these values are \textit{not} $0$ and $1$ (or $-1$ and $1$).  In fact, the actual values of the weights are not important: see Section \ref{subsubsec-weights}.  {\minirocket} does not use bitwise operations, and the input and convolution output are used at full precision.

The optimisations used in {\minirocket} are similar in motivation to several optimisations developed for convolutional neural networks, i.e., broadly speaking, to reduce the number of operations (especially multiplications) required to perform the convolution operation, e.g., \citep{liu_etal_2015,lavin_and_gray_2016,chollet_2017,mehta_etal_2018}.  In precomputing the product of the kernel weights and the input, and using those precomputed values to construct the convolution output (see Sections \ref{subsubsec-factoring-out} and \ref{subsubsec-all-kernels-at-once}), the optimisations used in {\minirocket} bear some resemblance to highly simplified versions of shift-based methods \citep{wu_etal_2018}, where conventional convolutional kernels are replaced by a combination of $1 \times 1$ convolutions and spatial shifts in the input, and lookup-based methods \citep{bagherinezhad_etal_2017}, where the convolution operation is performed via linear combinations of the precomputed convolution output for a small `dictionary' of kernels.

However, the optimisations used in {\minirocket} are much simpler than these methods.  {\minirocket} uses a fixed set of kernels, and uses the convolution output for these kernels directly, rather than through a learned linear combination, c.f., e.g., \citep{bagherinezhad_etal_2017,juefeixu_etal_2017}.  The optimisations arise as a natural result of using this fixed set of kernels, rather than being general-purpose optimisations.

Several things further distinguish {\minirocket} (and {\rocket}) from most approaches involving convolutional neural networks.  The features produced by the transform are all independent of each other (there is no hidden layer).  Neither the convolution output nor the pooled features are transformed through, e.g., a sigmoid function or rectified linear unit (ReLU).  As such, the classifier learns a direct linear function of the features produced by the transform.  {\minirocket} is also distinguished by its use of dilation (similar to using many different dilations in a single convolutional layer, with dilations taking any integer value not just powers of two), and PPV.

\section{Method} \label{sec-method}

{\minirocket} involves making certain key changes in order to remove almost all randomness from {\rocket} (Section \ref{subsec-removing-randomness}), and exploiting these changes in order to dramatically speed up the transform (Section \ref{subsec-optimising-the-transform}).  In tuning kernel length, weights, bias, etc., we have restricted ourselves to the same 40 `development' datasets as used in \citep{dempster_etal_2020}, with the same aim of avoiding overfitting the entire UCR archive.  (Note, however, that it is not necessarily the aim of {\minirocket} to maximise accuracy \textit{per se}, but rather to balance accuracy with parameter choices which remove randomness and are conducive to optimising the transform.)  The procedures for setting the parameter values and performing the transform are set out in \texttt{\ref{pseudo-fit}} and \texttt{\ref{pseudo-transform}} in Appendix \ref{sec-appendix-pseudocode}.

As for {\rocket}, we implement {\minirocket} in Python, compiled via Numba \citep{lam_etal_2015}.  We use a ridge regression classifier from scikit-learn \citep{pedregosa_etal_2011}, and logistic regression implemented using PyTorch \citep{paszke_etal_2019}.  Our code is available at: \url{https://github.com/angus924/minirocket}.

\subsection{Removing Randomness} \label{subsec-removing-randomness}

\begin{table}
  \centering
  \caption{Summary of changes from {\rocket} to {\minirocket}.}
  \begin{tabular}{ccc}
    \toprule
    & {\rocket} & {\minirocket} \\
    \midrule
    length & $\{7, 9, 11\}$ & 9 \\
    weights & $\mathcal{N}(0, 1)$ & $\{-1, 2\}$ \\
    bias & $\mathcal{U}(-1, 1)$ & from convolution output \\
    dilation & random & fixed (rel. to input length) \\
    padding & random & fixed \\
    features & PPV + max & PPV \\
    num. features & 20K & 10K \\
    \bottomrule
  \end{tabular}
  \label{table-changes-rocket-to-minirocket}
\end{table}

{\rocket} uses kernels with lengths selected randomly from $\{7,9,11\}$, weights drawn from $\mathcal{N}(0, 1)$, bias terms drawn from $\mathcal{U}(-1, 1)$, random dilations, and random paddings.  Two features, PPV and max, are computed per kernel, for a total of $20{,}000$ features.  {\minirocket} is characterised by a number of key changes to the kernels in terms of length, weights, bias, dilation, and padding, as well as resulting changes to the features, as summarised in Table \ref{table-changes-rocket-to-minirocket}.

\subsubsection{Length}

{\minirocket} uses kernels of length 9, with weights restricted to two values, building on the observation in \citep{dempster_etal_2020} that weights drawn from $\{-1, 0, 1\}$ produce similar accuracy to weights drawn from $\mathcal{N}(0, 1)$.

In order to maximise computational efficiency, the set of kernels should be as small as possible: see Section \ref{subsubsec-reusing-output}.  The set of possible two-valued kernels grows exponentially with length.  There are $2^{3} = 8$ possible kernels of length 3, but $2^{15} = 32{,}768$ possible kernels of length 15.  With more than two values, the set of possible kernels grows even faster with length.  For example, there are $3^{15} \approx 14\text{ million}$ possible three-valued kernels of length 15.

There are $2^{9} = 512$ possible two-valued kernels of length 9.  {\minirocket} uses a subset of 84 of these kernels, a subset which balances accuracy with the computational advantages of using a small number of kernels: see Section \ref{subsubsec-sensitivity-kernels}.  (A length of 9 is also consistent with the average length used in {\rocket}.)

\subsubsection{Weights} \label{subsubsec-weights}

Kernels with weights restricted to two values, $\alpha$ and $\beta$, can be characterised in terms of the number of weights with the value $\beta$ (or, equivalently, the number of weights with the value $\alpha$).  In this sense, the full set of two-valued kernels of length 9 includes the subset of kernels with 1 value of $\beta$ (e.g., $[\alpha,\alpha,\alpha,\alpha,\alpha,\alpha,\alpha,\alpha,\beta]$), the subset of kernels with 2 values of $\beta$ (e.g., $[\alpha,\alpha,\alpha,\alpha,\alpha,\alpha,\alpha,\beta,\beta]$), and so on.  {\minirocket} uses the subset kernels with 3 values of $\beta$:
\begin{gather*}
  [\alpha,\alpha,\alpha,\alpha,\alpha,\alpha,\beta,\beta,\beta] \\
  [\alpha,\alpha,\alpha,\alpha,\alpha,\beta,\alpha,\beta,\beta] \\
  [\alpha,\alpha,\alpha,\alpha,\beta,\alpha,\alpha,\beta,\beta] \\
  ...
\end{gather*}

For {\minirocket}, we set $\alpha = -1$ and $\beta = 2$.  However, the choice of $\alpha$ and $\beta$ is arbitrary, in the sense that the scale of these values is unimportant.  For an input time series, $X$, kernel, $W$, and bias, $b$, PPV is given by $\text{PPV}(X * W - b) = \frac{1}{n} \sum [X * W - b > 0]$ or, equivalently, $\text{PPV}(X * W) = \frac{1}{n} \sum [X * W > b]$, where `$*$' denotes convolution, and $[X \in a]$ denotes the indicator function.  As such, computing PPV is essentially equivalent to computing the empirical cumulative distribution function.  Accordingly, the scale of the weights is unimportant, because bias values are drawn from the convolution output, $X * W$ (see Section \ref{subsubsec-bias}), and so by definition match the scale of the weights and the scale of the input.  (Hence, in contrast to {\rocket}, it is not necessary to normalise the input.)

It is only important that the sum of the weights should be zero or, equivalently, that $\beta = -2 \alpha$.  Otherwise, the values of $\alpha$ and $\beta$ are not important.  This constraint---that the weights sum to zero---ensures that the kernels are only sensitive to the relative magnitude of the values in the input, i.e., that the convolution output is invariant to the addition or subtraction of any constant value to the input, i.e., $X * W = (X \pm c) * W$.

As PPV is bounded between 0 and 1, in computing PPV for a given kernel, $W$, we get an equivalent feature for the inverted kernel, $-W$, `for free': see Section \ref{subsubsec-computing-ppv}.  Accordingly, there is no need to use both the set of kernels with weights $\alpha = -1$ and $\beta = 2$, and the corresponding inverted set of kernels with weights $\alpha = 1$ and $\beta = -2$, as we get these inverted kernels `for free'.

The set of 84 kernels of length 9 with three weights with the value $\beta = 2$, and six weights with the value $\alpha = -1$, has the desirable properties of being a relatively small, fixed set of kernels---conducive to the optimisations pursued in Section \ref{subsec-optimising-the-transform}---and producing high classification accuracy.  However, we stress that there is not necessarily anything `special' about this set of kernels.  Other subsets of kernels of length 9, and kernels of other lengths, produce similar accuracy: see Section \ref{subsubsec-sensitivity-kernels}.  This is in addition to the observations in \citep{dempster_etal_2020}, i.e., that kernels (of various lengths) with weights drawn from $\mathcal{N}(0, 1)$, or from $\{-1, 0, 1\}$, are also effective.

\subsubsection{Bias} \label{subsubsec-bias}

Bias values are drawn from the convolution output, and are used to compute PPV as set out above in Section \ref{subsubsec-weights}.  By default, for a given kernel/dilation combination, bias values are drawn from the quantiles of the convolution output for a single, randomly-selected training example.  For a given kernel, $W$, and dilation, $d$, we compute the convolution output for a randomly-selected training example, $X$, i.e., $W_{d} * X$.  We take, e.g., the $[0.25, 0.5, 0.75]$ quantiles from $W_{d} * X$ as bias values, to be used in computing PPV.  We use a low-discrepancy sequence to assign quantiles to different kernel/dilation combinations \citep{schretter_etal_2016}.

The selection of training examples for the purpose of sampling bias values is the only stochastic element of {\minirocket}.  Further, while the choice of training example is random, in drawing bias values from the convolution output, we are selecting values produced by an otherwise entirely deterministic procedure.  This is why we characterise {\minirocket} as `minimally random'.

For the deterministic variant of {\minirocket}, bias values are drawn from the convolution output for the entire training set, rather than a single, randomly-selected training example.  This is the only substantive difference between the default and deterministic variants, and the difference in accuracy between the two variants is negligible: see Section \ref{subsec-ucr-archive}.

The advantage of using the entire training set is an entirely deterministic transform, for applications where this is desirable.  However, this comes at additional computational cost, which is unlikely to be practical for larger datasets.  Crucially, however, it demonstrates that the accuracy of {\rocket} is achievable using an entirely deterministic transform.  In practice, using a single, randomly-selected training example has little impact in terms of accuracy.

A variant of {\rocket} using the same method for sampling bias values as {\minirocket} is slightly more accurate than default {\rocket} but, overall, the difference is relatively minor: see Section \ref{subsec-ucr-archive}.

\subsubsection{Dilation} \label{subsubsec-dilation}

Dilation is used to `spread' a kernel over the input.  For dilation, $d$, a given kernel is convolved with every $d^{\text{th}}$ element of the input \citep{yu_and_koltun_2016,dempster_etal_2020}.  Each kernel is assigned the same fixed set of dilations, adjusted to the length of the input time series.  We specify dilations in the range $D = \{\lfloor 2^{0} \rfloor, ..., \lfloor 2^{\text{max}} \rfloor\}$, where the exponents are uniformly spaced between 0 and $\text{max} = \log_2 ( l_{\text{input}} - 1 ) / ( l_{\text{kernel}} - 1 )$, where $l_{\text{input}}$ is input length and $l_{\text{kernel}}$ is kernel length (i.e., 9), such that the maximum effective length of a kernel, including dilation, is the length of the input time series.  The count of each unique integer dilation value in $D$ determines the number of features to be computed per dilation (scaled according to the total number of features), ensuring that, as in {\rocket}, exponentially more features are computed for smaller dilations.

As time series length increases, the number of possible dilation values increases.  This means that, for a fixed number of features, the number of features computed per dilation decreases (unless constrained in some way), making the transform less efficient: see Section \ref{subsubsec-reusing-output}.  Hence, by default, we limit the maximum number of dilations per kernel to 32.  While technically an additional hyperparameter, this has little effect on accuracy (see Section \ref{subsubsec-sensitivity-dilation}), and is intended to be kept at its default value.

\subsubsection{Padding} \label{subsubsec-padding}

Padding is alternated for each kernel/dilation combination such that, overall, half the kernel/dilation combinations use padding, and half do not.  As for {\rocket}, {\minirocket} uses standard zero padding.  In effect, zeros are added to the start and end of each input time series such that the convolution operation begins with the `middle' element of the kernel centered on the first element of the time series, and ends with the `middle' element of the kernel centered on the last element of the time series \citep{goodfellow_etal_2016}.

\subsubsection{Features} \label{subsubsec-features}

Given the other changes, there is no longer any benefit in terms of accuracy in using global max pooling in addition to PPV: see Section \ref{subsubsec-sensitivity-features}.  Accordingly, {\minirocket} `drops' global max pooling and uses only PPV.

We do not replace global max pooling with additional PPV features.  As for {\rocket}, the number of features represents a tradeoff between accuracy and computational expense.  {\minirocket} with $10{,}000$ features already matches {\rocket} in terms of accuracy, and there is little or no benefit in terms of accuracy to increasing the number of features beyond $10{,}000$: see Section \ref{subsubsec-sensitivity-num-features}.

Accordingly, by default, {\minirocket} uses $10{,}000$ features (or, more precisely, the nearest multiple of 84---the number of kernels---less than $10{,}000$, i.e., $9{,}996$).  While technically a hyperparameter, this is intended to be kept at its default value.

\subsection{Optimising the Transform} \label{subsec-optimising-the-transform}

{\minirocket} takes advantage of the properties of the small, fixed set of two-valued kernels, and of PPV, to significantly speed up the transform through four key optimisations:

\begin{enumerate}
  \item computing PPV for $W$ and $-W$ at the same time;
  \item reusing the convolution output to compute multiple features;
  \item avoiding multiplications in the convolution operation; and
  \item for each dilation, computing all kernels (almost) `at once'.
\end{enumerate}

\subsubsection{Computing PPV for $W$ and $-W$ at the Same Time} \label{subsubsec-computing-ppv}

\begin{figure}
\centering
\includegraphics[width=\linewidth]{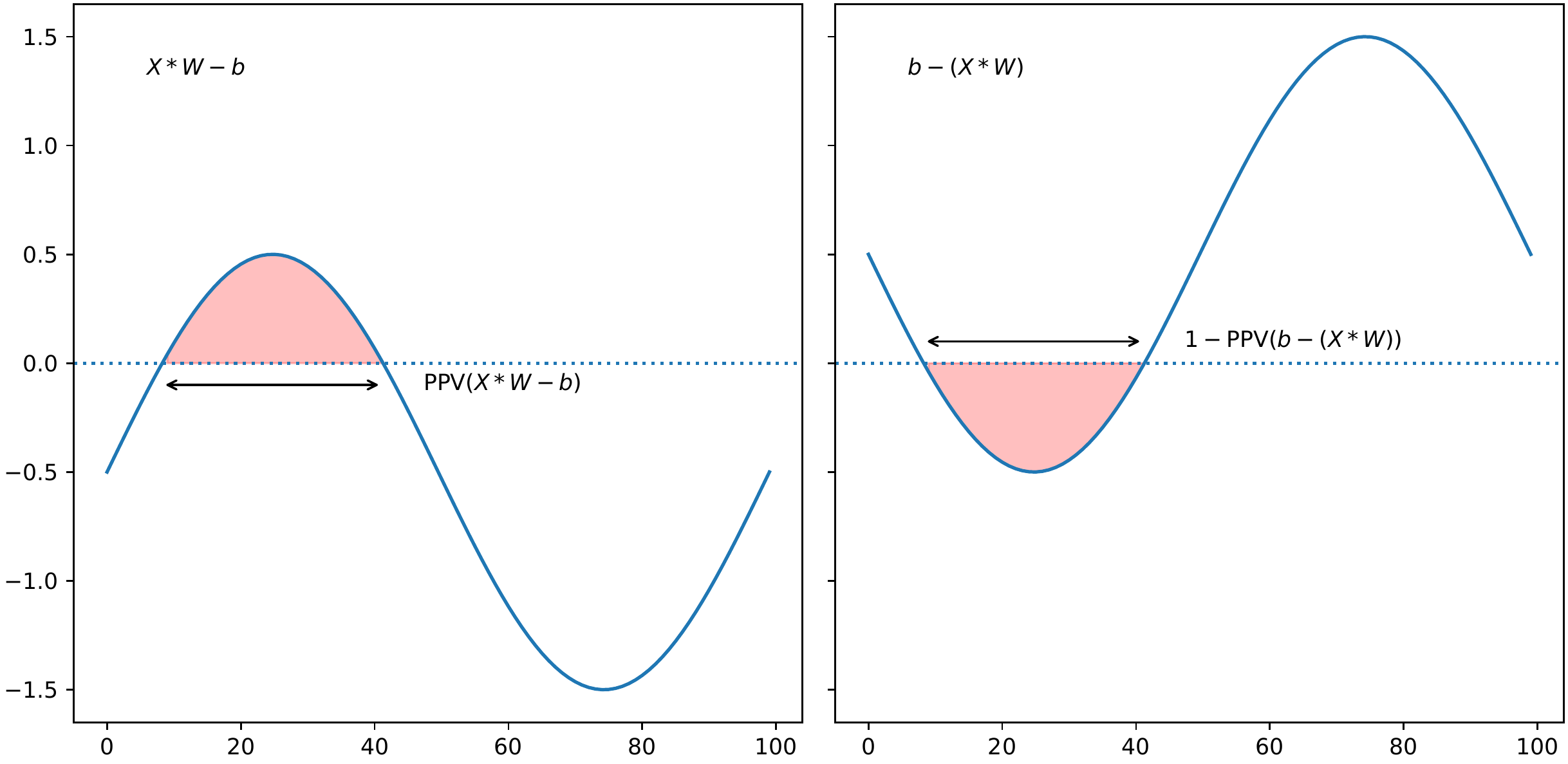}
\caption{Illustration of $\text{{\normalfont PPV}}(X * W - b) = 1 - \text{{\normalfont PPV}}(b - (X * W))$.}
\Description[PPV for a given kernel, W, is equivalent to a very similar feature for the inverted kernel, negative W]{A toy example showing that PPV for a given kernel, W, is equivalent to a very similar feature for the inverted kernel, negative W.  Two plots: the plot on the left shows the convolution output for input X and kernel W minus bias b, and the resulting proportion of positive values, or PPV; the plot on the right shows the convolution output for the same input X and the inverted kernel negative W plus bias b, and the proportion of negative values, or PNV.  PNV for the inverted kernel negative W is shown to be the same as PPV for the kernel W.}
\label{fig-diagram-ppv-inverse}
\end{figure}

For $C = X * W - b$, PPV is given by $\text{PPV}(C) = \frac{1}{n} \sum [c > 0].$  PPV is bounded between 0 and 1.  By definition, the proportion of negative values (or PNV) is the complement of PPV, i.e., $1 - \text{PPV}(X * W - b) = \text{PNV}(X * W - b).$  That is, by computing PPV, we also implicitly compute PNV and vice versa.  In this sense, PPV and PNV are equivalent.

Further, the convolution operation is associative, such that $X * -W = -(X * W)$.  Accordingly, by computing PPV for a given kernel, $W$, we unavoidably also compute an equivalent feature (i.e., PNV) for $-W$, that is, $\text{PPV}(X * W - b) = 1 - \text{PPV}(b - (X * W)).$  This relationship is illustrated in Figure \ref{fig-diagram-ppv-inverse}.  This means that, for the purposes of PPV, it is unnecessary to compute both $X * W$ and $X * -W$.  In fact, it would be redundant to do so.

This means that, in practice, for a given set of kernels where each kernel, $W$, is matched by a corresponding inverted kernel, $-W$, we only need to perform the convolution operation for $W$, i.e., for half of the kernels.  We get $-W$ `for free'.  Accordingly, as set out in Section \ref{subsubsec-weights}, {\minirocket} only uses a set of kernels with weights $\alpha = -1$ and $\beta = 2$, as it is unnecessary to also use the corresponding set of inverted kernels with weights $\alpha = 1$ and $\beta = -2$.

\subsubsection{Reusing the Convolution Output} \label{subsubsec-reusing-output}

For {\minirocket}, the same kernel/dilation combination is used to compute multiple features, at least for smaller dilations (exponentially fewer features are computed for larger dilations: see Section \ref{subsubsec-dilation}).

For a given kernel, $W$, and dilation, $d$, we compute $C = X * W_{d}$ and then reuse the convolution output, $C$, to compute multiple features, i.e., for multiple different bias values.  This has the effect that multiple features are computed with the computational cost of a single convolution operation, plus the much lower cost of computing PPV for each bias value.

\subsubsection{Avoiding Multiplications} \label{subsubsec-factoring-out}

Restricting the kernel weights to two values allows us to, in effect, `factor out' the multiplications from the convolution operation, and to perform the convolution operation using only addition.

For input time series $X = [x_{0}, x_{1}, ..., x_{n - 1}]$, and kernel $W = [w_{0}, w_{1}, ..., w_{m - 1}]$, with dilation, $d$, the convolution operation can be formulated as:
$$
X * W_{d} = \sum_{j=0}^{m - 1} x_{i - (\lfloor \frac{m}{2} \rfloor \cdot d) + (j \cdot d)} \cdot w_{j}, \forall i \in \{0, 1, ..., n - 1\}.
$$

Equivalently, the convolution operation can be thought of as the column sums of a matrix, $\boldsymbol{\hat{C}}$, where each row corresponds to the input time series multiplied by the appropriate kernel weight, and the alignment of the rows corresponds to dilation (values of 0 in $\boldsymbol{\hat{C}}$ represent zero padding), e.g.:
$$
\hat{\boldsymbol{C}} =
\begin{bmatrix}
  0 & 0 & 0 & 0 & w_{0} x_{0} & \cdots \\
  0 & 0 & 0 & w_{1} x_{0} & w_{1} x_{1} & \cdots \\
  0 & 0 & w_{2} x_{0} & w_{2} x_{1} & w_{2} x_{2} & \cdots \\
  \vdots & \vdots & \vdots & \vdots & \vdots & \ddots \\
  w_{m - 1} x_{4} & w_{m - 1} x_{5} & w_{m - 1} x_{6} & w_{m - 1} x_{7} & w_{m - 1} x_{8} & \cdots
\end{bmatrix}
$$

The result of the convolution operation is given by the column sums of $\boldsymbol{\hat{C}}$, i.e., $C = X * W = \boldsymbol{1}^{\top}\boldsymbol{\hat{C}}$, where $\boldsymbol{1}$ is a vector, $[1,1,...,1]^{\top}$, of length $n$.

Where the weights of the kernels are restricted to two values, $\alpha$ and $\beta$, we can `factor out' the multiplications by precomputing $A = \alpha X$ and $B = \beta X$ and then, for a given kernel, e.g., $W = [\alpha, \beta, \alpha, ..., \alpha]$, completing the convolution operation by summation using $A = [a_{0}, a_{1}, ..., a_{n - 1}]$ and $B = [b_{0}, b_{1}, ..., b_{n - 1}]$:
$$
\hat{\boldsymbol{C}} =
\begin{bmatrix}
  0 & 0 & 0 & 0 & a_{0} & \cdots & a_{n-5} \\
  0 & 0 & 0 & b_{0} & b_{1} & \cdots & b_{n-4} \\
  0 & 0 & a_{0} & a_{1} & a_{2} & \cdots & b_{n-3} \\
  \vdots & \vdots & \vdots & \vdots & \vdots & \ddots & \vdots \\
  a_{4} & a_{5} & a_{6} & a_{7} & a_{8} & \cdots & 0
\end{bmatrix}
$$

In other words, it is only necessary to compute $\alpha X$ and $\beta X$ once for each input time series, and then reuse the results to complete the convolution operation for each kernel by addition.

\subsubsection{Computing All the Kernels (Almost) `At Once'} \label{subsubsec-all-kernels-at-once}

We can take further advantage of using only two values for the kernel weights in order to perform most of the computation required for all 84 kernels `at once' for each dilation value.  More precisely, as {\minirocket} uses kernels with six weights of one value, and three weights of another value, we can perform $\frac{6}{9} = \frac{2}{3}$ of the computation for all 84 kernels `at once' for a given dilation.

This is possible by treating all kernel weights as $\alpha = -1$, precomputing convolution output, $C_{\alpha}$, and later adjusting $C_{\alpha}$ for each kernel.  Per Section \ref{subsubsec-factoring-out}, $C_{\alpha}$ can be thought of as the column sums of a matrix with 9 rows, where each row corresponds to $\alpha X = -X$, aligned according to dilation.  For example, for a dilation of 1:
$$
\hat{\boldsymbol{C}}_{\alpha} =
\begin{bmatrix}
  0 & 0 & 0 & 0 & -x_{0} & \cdots & -x_{n-5} \\
  0 & 0 & 0 & -x_{0} & -x_{1} & \cdots & -x_{n-4} \\
  0 & 0 & -x_{0} & -x_{1} & -x_{2} & \cdots & -x_{n-3} \\
  \vdots & \vdots & \vdots & \vdots & \vdots & \ddots & \vdots \\
  -x_{4} & -x_{5} & -x_{6} & -x_{7} & -x_{8} & \cdots & 0
\end{bmatrix}
$$

For {\minirocket}, the kernel weights are $\alpha = -1$ and $\beta = 2$.  Let $\gamma = 3$, noting that $2 = -1 + 3$.  As for $\hat{\boldsymbol{C}}_{\alpha}$, we then form $\hat{\boldsymbol{C}}_{\gamma}$, where each row corresponds to $\gamma X = 3X$, aligned according to dilation.  For each kernel, $C_{\gamma}$ is equivalent to the column sums of those rows in $\hat{\boldsymbol{C}}_{\gamma}$ corresponding to the position of the $\beta$ weights in the given kernel.  For example, for kernel $W = [\beta, \alpha, \beta, \alpha, \beta, \alpha, \alpha, \alpha, \alpha]$:
$$
\hat{\boldsymbol{C}}_{\gamma}^{(W)} =
\begin{bmatrix}
  0 & 0 & 0 & 0 & 3x_{0} & \cdots & 3x_{n-5} \\
  0 & 0 & 3x_{0} & 3x_{1} & 3x_{2} & \cdots & 3x_{n-3} \\
  3x_{0} & 3x_{1} & 3x_{2} & 3x_{3} & 3x_{4} & \cdots & 3x_{n-1}
\end{bmatrix}
$$

The final convolution output for a given kernel is then given by $C = C_{\alpha} + C_{\gamma}$.  In other words, we can reuse $C_{\alpha}$, computed once for a given dilation, to compute the convolution output for all 84 kernels for that dilation.  For each kernel, computing $C$ only involves adding $C_{\gamma}$ to $C_{\alpha}$.  In performing the convolution operation in this way, we only have to compute $C_{\gamma}$ for each kernel, i.e., $\frac{1}{3}$ of the computation otherwise required.

\subsection{Classifiers}

Like {\rocket}, {\minirocket} is a transform, producing features which are then used to train a linear classifier.  We use the same classifiers as {\rocket} to learn the mapping from the features to the classes, i.e., a ridge regression classifier or, for larger datasets, logistic regression trained using Adam \citep{kingma_and_ba_2015}.  As for {\rocket}, we suggest switching from the ridge regression classifier to logistic regression when there are more training examples than features, i.e., when there are more than approximately $10{,}000$ training examples.

\subsection{Complexity}

Fundamentally, the scalability of {\minirocket} remains the same as for {\rocket}: linear in the number of kernels/features ($k$), the number of examples ($n$), and time series length ($l_{\text{input}}$) or, formally, $O(k \cdot n \cdot l_{\text{input}})$.  While {\minirocket} uses a smaller number of kernel/dilation combinations, and computes multiple features for each kernel/dilation combination, complexity is still proportional to the number of kernels/features.  Similarly, while {\minirocket} `factors out' multiplications from the convolution operation, the number of addition operations is still proportional to the number of kernels and time series length, and while {\minirocket} performs the majority of the computation required for all 84 kernels `at once', the remaining computation is still proportional to the number of kernels/features.  However, within this broad class of complexity, the various optimisations pursued in Section \ref{subsec-optimising-the-transform} make {\minirocket} significantly faster in practice.

\subsection{Memory}

Compared to {\rocket} (which does not store any intermediate values), {\minirocket} temporarily stores up to 13 additional vectors, namely, $A = -X$, $G = \gamma X = 3X$ (plus 9 variants of $G$ pre-aligned for the given dilation), $C_{\alpha}$, and $C$: see Sections \ref{subsubsec-factoring-out} and \ref{subsubsec-all-kernels-at-once}.  This is equivalent to storing 13 additional copies of a single input time series (approx. $1{,}000{,}000 \times 4 \times 13 = 52 \text{MB}$ for time series of length 1 million), which should be negligible in almost all cases.

When transforming the training set, the deterministic variant stores the convolution output for a given kernel/dilation combination for the entire training set, which is equivalent to storing one additional copy of the entire training set.  This is impractical for larger datasets, which is why it is avoided by default.

\section{Experiments} \label{sec-experiments}

We evaluate {\minirocket} on the datasets in the UCR archive (Section \ref{subsec-ucr-archive}), showing that, on average, {\minirocket} is marginally more accurate than {\rocket}, and not significantly less accurate than the most accurate current methods for time series classification.  We demonstrate the speed and scalability of {\minirocket} in terms of both training set size and time series length (Section \ref{subsec-scalability}), showing that {\minirocket} is up to $75$ times faster than {\rocket} on larger datasets.  We also explore the effect of key parameters in relation to kernel length, bias, output features, and dilation (Section \ref{subsec-sensitivity-analysis}).

\subsection{UCR Archive} \label{subsec-ucr-archive}

We evaluate {\minirocket} on the datasets in the UCR archive \citep{dau_etal_2019}.  We compare {\minirocket} against the most accurate current methods for time series classification, namely, HIVE-COTE/TDE (representative of HIVE-COTE and its variants), TS-CHIEF, InceptionTime, and {\rocket}, as well as TDE, CIF, cBOSS and Proximity Forest.

For consistency and direct comparability with the most recent published results for other state-of-the-art methods \citep{bagnall_etal_2020,middlehurst_etal_2020a,middlehurst_etal_2020b}, we evaluate {\minirocket} on 30 resamples of 109 datasets from the archive.  We use the same 30 resamples (including the default training/test split) as in \citep{bagnall_etal_2020,middlehurst_etal_2020a,middlehurst_etal_2020b}.  (Full results are available in the accompanying repository.)

Figure \ref{fig-rank-ucr109} on page \pageref{fig-rank-ucr109} shows the mean rank of {\minirocket} versus the other state-of-the-art methods.  Methods for which the pairwise difference in accuracy is not statistically significant, per a Wilcoxon signed-rank test with Holm correction (as a post hoc test to the Friedman test), are connected with a black line \citep{demsar_2006,garcia_and_herrera_2008,benavoli_etal_2016}.

{\minirocket} is, on average, marginally more accurate than {\rocket}, and somewhat less accurate than the most accurate current methods, namely TS-CHIEF and HIVE-COTE/TDE, although the differences in accuracy are not statistically significant.  However, as noted in Section \ref{sec-introduction}, the total compute time for {\minirocket} on these datasets is a tiny fraction of the total compute time required by the other methods (even {\rocket}, which is already considerably faster than even the fastest of the other methods).

\paragraph{{\minirocket} versus {\rocket}.}

\begin{figure}
\centering
\includegraphics[width=0.65\linewidth]{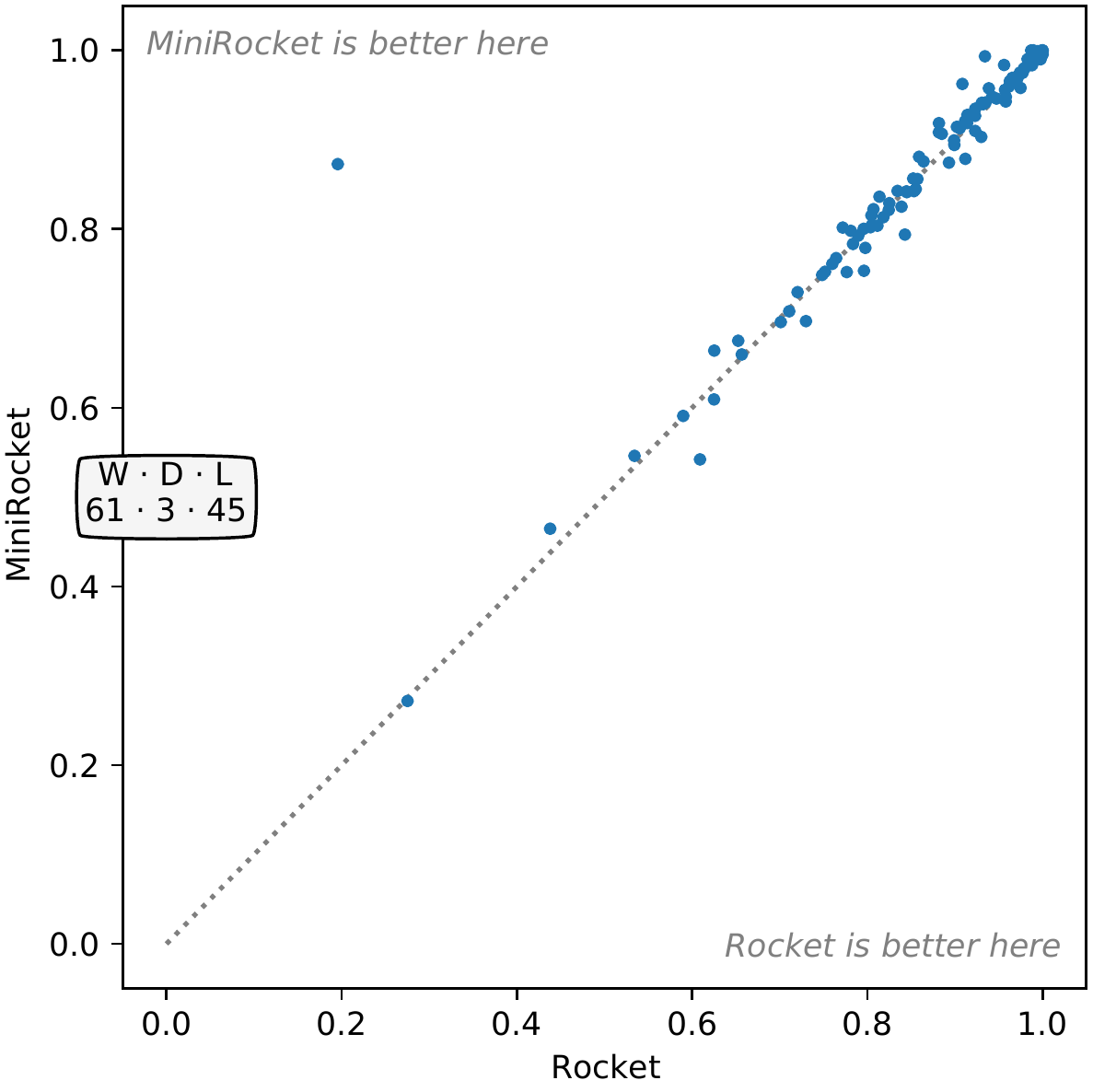}
\caption{Pairwise accuracy of {\minirocket} versus {\rocket}.}
\Description[Apricot is more accurate than Rocket on 61 of 109 datasets]{Scatter plot showing the accuracy of Apricot against the accuracy of Rocket for 109 datasets from the UCR archive.  Apricot is more accurate on 61 datasets, as accurate on 3 datasets, and less accurate on 45 datasets.  The accuracy of Apricot and Rocket is similar for most datasets.  For one dataset, PigAirWayPressure, Apricot is considerably more accurate than Rocket.}
\label{fig-pairwise-minirocket-vs-rocket}
\end{figure}

Figure \ref{fig-pairwise-minirocket-vs-rocket} shows the pairwise accuracy of {\minirocket} versus {\rocket} for the same 109 datasets.  Overall, {\minirocket} and {\rocket} achieve very similar accuracy.  {\minirocket} is more accurate than {\rocket} on 61 datasets, and less accurate on 45 datasets, but the differences in accuracy are mostly small.  The large difference in accuracy between {\minirocket} and {\rocket} on one dataset, \textit{PigAirwayPressure}, appears to be due to the way the bias values are sampled.  We also evaluated a variant of {\rocket} which uses the same method of sampling bias values as {\minirocket}.  Overall, this variant is slightly more accurate than default {\rocket}, but the difference is relatively minor, with a win/draw/loss of 50/6/53 against {\minirocket}.

\paragraph{Deterministic variant.}

\begin{figure}
\centering
\includegraphics[width=0.65\linewidth]{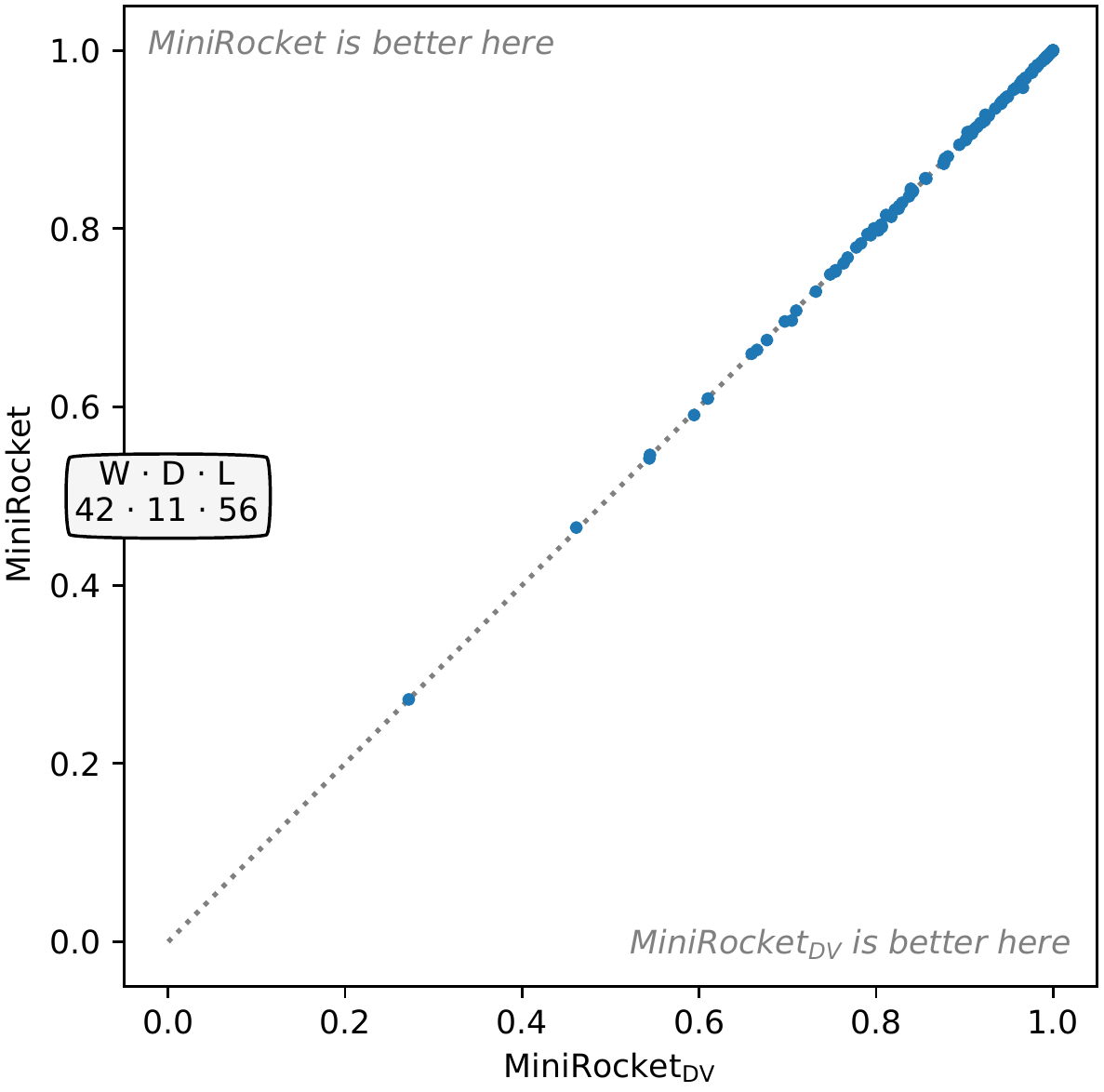}
\caption{Pairwise accuracy of default {\minirocket} versus the deterministic variant.}
\Description[Apricot is more accurate than the deterministic variant of Apricot on 42 of 109 datasets]{Scatter plot showing the accuracy of Apricot against the accuracy of the deterministic variant of Apricot for 109 datasets from the UCR archive.  Apricot is more accurate on 42 datasets, as accurate on 11 datasets, and less accurate on 56 datasets.  The accuracy of Apricot and the deterministic variant is extremely similar on all datasets, with only very minor, almost imperceptible, differences.}
\label{fig-pairwise-minirocket-vs-minirocket-dv}
\end{figure}

Figure \ref{fig-pairwise-minirocket-vs-minirocket-dv} shows the pairwise accuracy of default {\minirocket} vs the deterministic variant (or {\minirocket}$_{\text{DV}}$) for the same 109 datasets.  Overall, the deterministic variant produces essentially the same accuracy as the default variant.

\subsection{Scalability} \label{subsec-scalability}

\subsubsection{Training Set Size}

\begin{figure*}
\centering
{\includegraphics[width=0.70\linewidth]{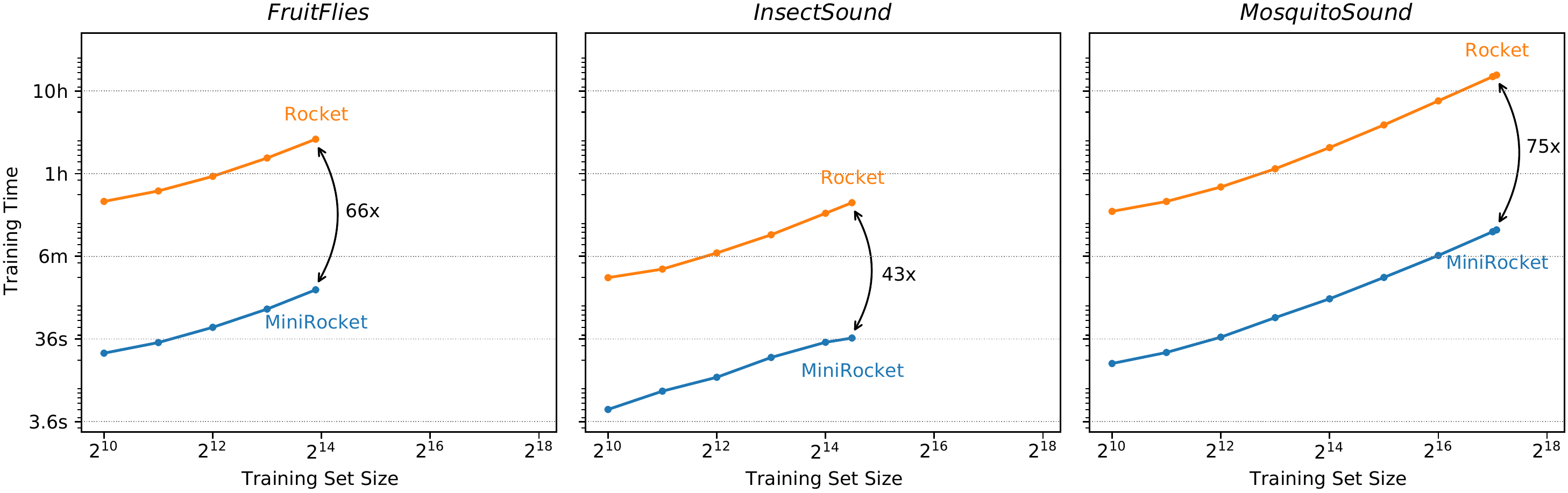}\phantomsubcaption}
\hfill
{\includegraphics[width=0.230\linewidth]{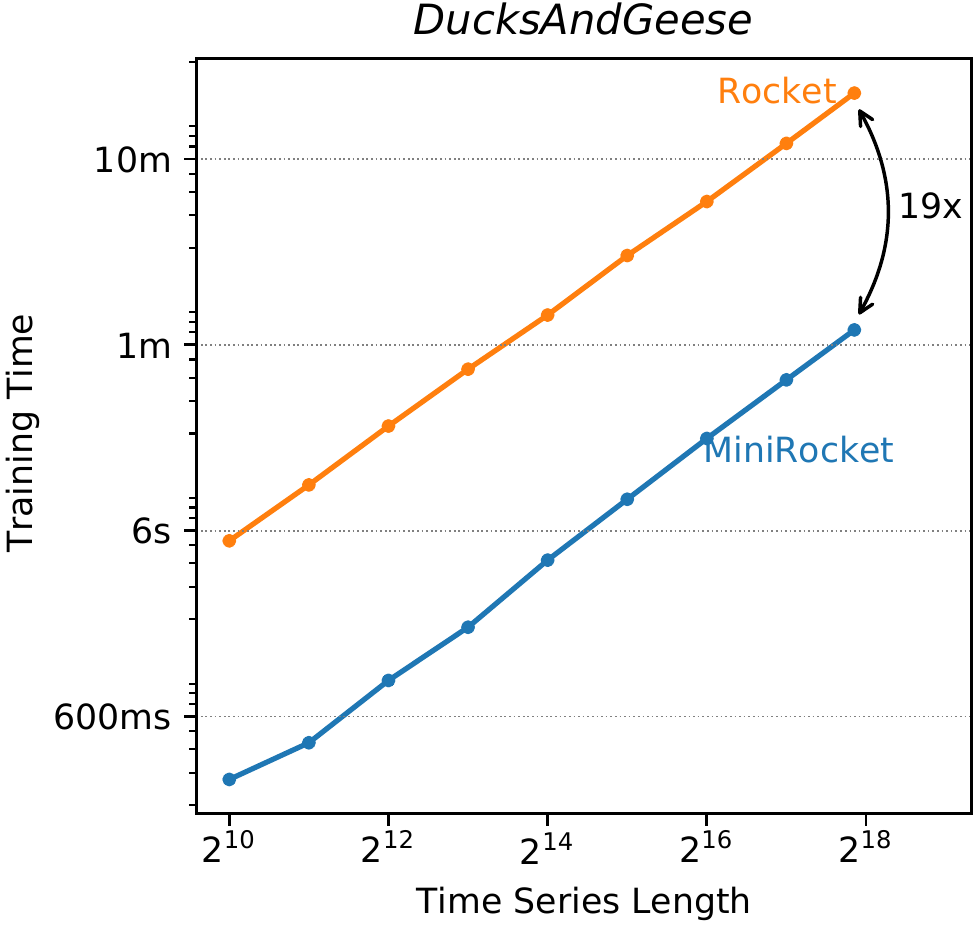}\phantomsubcaption}
\caption{Training time versus (left) training set size and (right) time series length.}
\Description[Apricot is much faster than Rocket on large datasets in terms of both training set size and time series length]{Three line plots showing training set size against total training time (transform plus classifier training), and one line plot showing time series length against total training time, for both Apricot and Rocket.  In terms of training set size, Apricot is 66 times faster for the FruitFlies dataset, 43 times faster for the InsectSound dataset, and 75 times faster for the MosquitoSound dataset.  In terms of time series length, Apricot is 19 times faster for the DucksAndGeese dataset.}
\label{fig-scalability}
\end{figure*}

\begin{table}
  \centering
  \caption{Accuracy and total training time.}
  \begin{tabular}{cccrr}
    \toprule
    & \multicolumn{2}{c}{Accuracy} & \multicolumn{2}{c}{Training Time} \\
    \cmidrule(lr){2-3} \cmidrule(lr){4-5}
    & {\rocket} & {\minirocket} & {\rocket} & {\minirocket} \\
    \midrule
    \textit{Fruit} & 0.9491 & 0.9568 & $2\text{h }36\text{m }40\text{s }$ & $2\text{m }22\text{s }$ \\
    \textit{Insect} & 0.7796 & 0.7639 & $26\text{m }44\text{s }$ & $37\text{s }$ \\
    \textit{Mosquito} & 0.8271 & 0.8165 & $15\text{h }34\text{m }58\text{s }$ & $12\text{m }32\text{s }$ \\
    \bottomrule
  \end{tabular}
  \label{table-scalability-training-set-size}
\end{table}

We demonstrate the speed and scalability of {\minirocket} in terms of training set size on the three largest datasets in the UCR archive, namely, \textit{MosquitoSound} ($139{,}780$ training examples, each of length $3{,}750$), \textit{InsectSound} ($25{,}000$ training examples, each of length $600$), and \textit{FruitFlies} ($17{,}259$ training examples, each of length $5{,}000$).  These recent additions are significantly larger than other datasets in the archive.

For this purpose, following \citep{dempster_etal_2020}, we integrate {\minirocket} (and {\rocket}) with logistic regression, trained using Adam.  Training details are provided in Appendix \ref{sec-appendix-training-details}.  The experiments were performed on the same cluster as noted in Section \ref{sec-introduction} and, again, both {\rocket} and {\minirocket} are restricted to a single CPU core.

Figure \ref{fig-scalability} shows training time vs training set size for {\minirocket} and {\rocket}.  Training time includes the transform for both validation and training sets, and classifier training.  Table \ref{table-scalability-training-set-size} shows test accuracy and total training time (for the full training set).

{\minirocket} is slightly more accurate on one of the datasets, and slightly less accurate on two of the datasets.  This is consistent with the small differences in accuracy observed on the other datasets in the UCR archive: see Section \ref{subsec-ucr-archive}.  However, {\minirocket} is considerably faster than {\rocket}: $43$ times faster on \textit{InsectSound}, $66$ times faster on \textit{FruitFlies}, and $75$ times faster on \textit{MosquitoSound}.

The accuracy of {\rocket} and {\minirocket} on the \textit{InsectSound} and \textit{MosquitoSound} datasets appears to be broadly comparable to reported results for other methods for these datasets or versions of these datasets \citep{chen_etal_2014,zhang_etal_2017,fanioudakis_etal_2018,flynn_and_bagnall_2019}, although some deep learning approaches are significantly more accurate on \textit{MosquitoSound} \citep{fanioudakis_etal_2018}.

\subsubsection{Time Series Length}

\begin{sloppypar}
  We demonstrate the scalability of {\minirocket} in terms of time series length on the dataset in the UCR archive with the longest time series, \textit{DucksAndGeese} ($50$ training examples, each of length $236{,}784$).  This recent addition has significantly longer time series than other datasets in the archive.
\end{sloppypar}

Figure \ref{fig-scalability} shows training time versus time series length for both {\rocket} and {\minirocket}.  Training time includes the transform and classifier training.  (With only 50 training examples, we use the ridge regression classifier.)

While both {\rocket} and {\minirocket} are linear in time series length, {\minirocket} is considerably faster for a given length.  With more training examples, we would expect the difference in training time to be considerably larger.  With only 50 training examples, the overhead of sampling bias values (which is unrelated to training set size) constitutes a significant proportion of the total training time for {\minirocket}.

\subsection{Sensitivity Analysis} \label{subsec-sensitivity-analysis}

We explore the effect of key parameter choices on accuracy:

\begin{itemize}
  \item kernel length;
  \item sampling bias from the convolution output versus $\mathcal{U}(-1, 1)$;
  \item using only PPV versus both PPV and global max pooling;
  \item the number of features; and
  \item limiting the maximum number of dilations per kernel.
\end{itemize}

We perform the analysis using the 40 `development' datasets (default training/test splits).  Results are mean results over 10 runs.

\subsubsection{Kernels} \label{subsubsec-sensitivity-kernels}

\begin{figure}
\centering
\includegraphics[width=\linewidth]{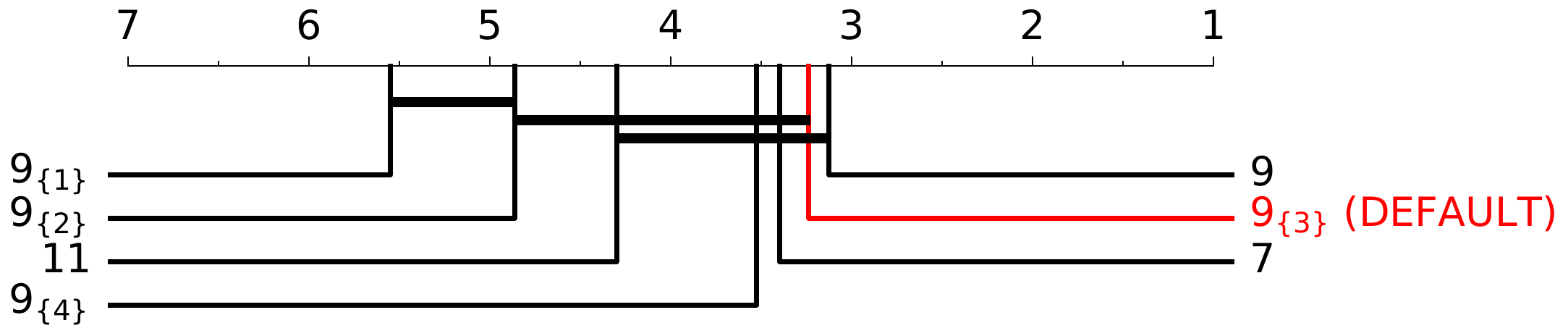}
\caption{Mean rank for different kernel lengths.}
\Description[The subset of kernels of length 9 having three weights of one value ranks just behind the full set of length 9]{The subset of kernels of length 9 having three weights of one value and six weights of another value ranks just behind the full set of kernels of length 9, and ahead of any other subset of length 9 and kernels of length 7 or 11.  The pairwise differences are (with one exception) not statistically significant.}
\label{fig-sensitivity-kal}
\end{figure}

Figure \ref{fig-sensitivity-kal} shows the effect of kernel length on accuracy.  For kernels of length 9, a subscript refers to a particular subset of kernels in the sense discussed in Section \ref{subsubsec-weights}.  (E.g., $9_{\{3\}}$ refers to kernels with three weights of one value, and six weights of another value.)  The total number of features is kept constant (to the nearest multiple of the number of kernels less than $10{,}000$: see Section \ref{subsubsec-features}), such that more features are computed per kernel for smaller sets of kernels and vice versa.

Kernels of length 9 are most accurate, but kernels of length 7 or 11 are not significantly less accurate.  This is consistent with the findings in \citep{dempster_etal_2020} in relation to {\rocket}.  The actual differences in accuracy between kernels of different lengths is very small.

Crucially, however, as noted in Section \ref{subsubsec-weights}, the $9_{\{3\}}$ subset is nearly as accurate as the full set of kernels of length 9.  This is a relatively small subset of kernels, and is particularly well suited to the optimisations pursued in Section \ref{subsec-optimising-the-transform}.

\subsubsection{Bias} \label{subsubsec-sensitivity-bias}

\begin{figure}
\centering
\includegraphics[width=\linewidth]{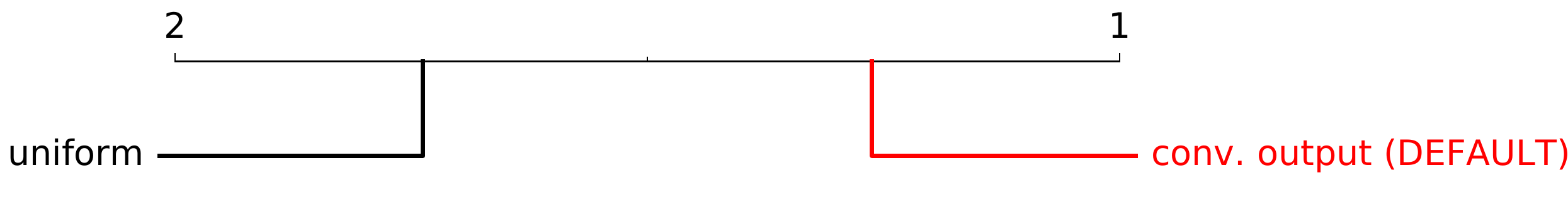}
\caption{Mean rank for bias sampled from the convolution output versus bias sampled from $\mathcal{U}(-1, 1)$.}
\Description[Bias sampled from the convolution output ranks well ahead of bias sampled uniformly]{Bias sampled from the convolution output ranks well ahead of bias sampled uniformly from negative 1 to 1, and the difference is statistically significant.}
\label{fig-sensitivity-bias}
\end{figure}

Figure \ref{fig-sensitivity-bias} shows the effect in terms of accuracy of sampling bias from the convolution output versus from $\mathcal{U}(-1, 1)$ as in {\rocket}.  {\minirocket} is significantly less accurate when sampling bias from $\mathcal{U}(-1,1)$.  The change to sampling bias from the convolution output is critical to matching the accuracy of {\rocket}.

\subsubsection{Features} \label{subsubsec-sensitivity-features}

\begin{figure}
\centering
\includegraphics[width=\linewidth]{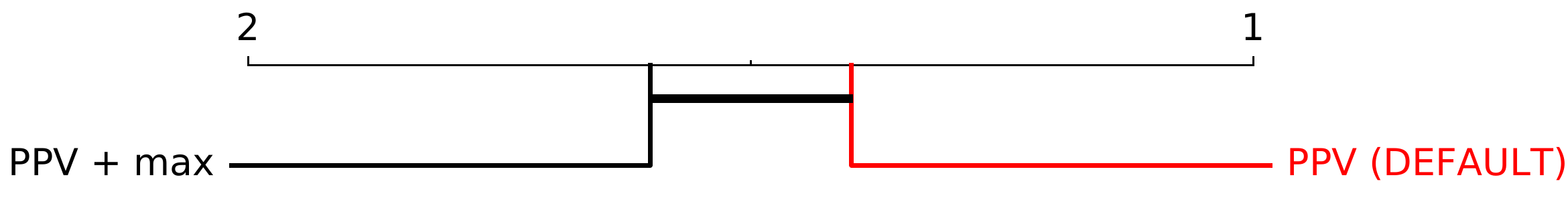}
\caption{Mean rank for PPV vs PPV and global max pooling.}
\Description[PPV by itself ranks ahead of the combination of PPV and global max pooling]{PPV by itself ranks ahead of the combination of PPV and global max pooling, but the difference is not statistically significant.}
\label{fig-sensitivity-ppv}
\end{figure}

Figure \ref{fig-sensitivity-ppv} shows the effect of using only PPV versus both PPV and global max pooling.  With the other changes to {\minirocket}---in particular, with the change to sampling bias from the convolution output---there is no advantage to using global max pooling in addition to PPV.  In fact, using global max pooling in addition to PPV is less accurate than just using PPV, although the difference is not statistically significant.

\subsubsection{Number of Features} \label{subsubsec-sensitivity-num-features}

\begin{figure}
\centering
\includegraphics[width=\linewidth]{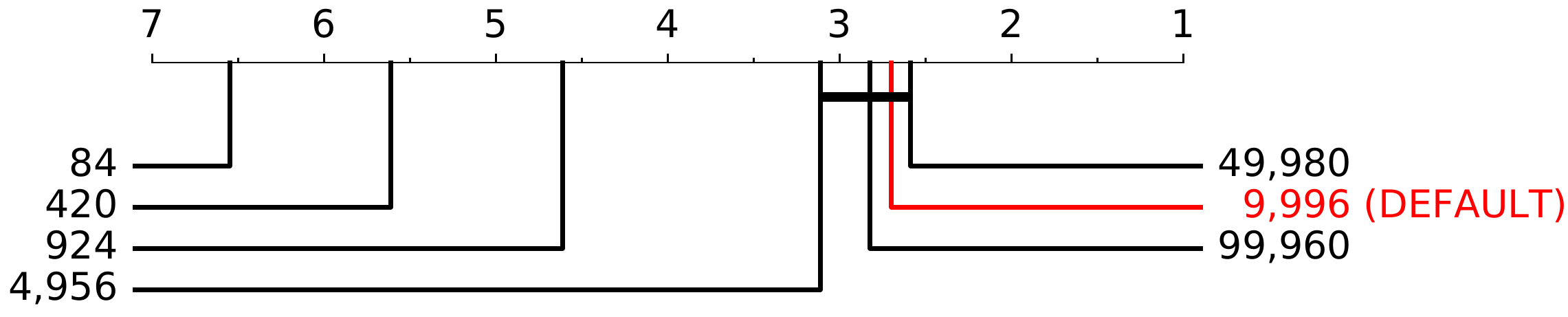}
\caption{Mean rank for different numbers of features.}
\Description[9,996 features, the default, ranks just behind 49,980 features, but ahead of any other number of features]{9,996 features, the default, ranks just behind 49,980 features, but ahead of 99,960 features and any smaller number of features.  9,996 features is in the same clique as 4,956, 99,960, and 49,980 features, that is, the pairwise differences are not statistically significant.}
\label{fig-sensitivity-num-features}
\end{figure}

Figure \ref{fig-sensitivity-num-features} shows the effect of different numbers of features between $84$ and $99{,}960$ (the nearest multiple of 84 less than 100, 500, $1{,}000$, ...).  Increasing the number of features noticeably increases accuracy up to approximately $10{,}000$ features.  There is little or no benefit to increasing the number of features beyond $10{,}000$, at least for shorter time series, because there is little benefit in computing PPV for many more than $l_{\text{input}}$ bias values for time series of length $l_{\text{input}}$ (more and more features will be the same).  For $49{,}980$ and $99{,}960$ features, we have endeavoured to avoid this limitation as much as possible by setting the maximum number of dilations per kernel to 119 (see Section \ref{subsubsec-dilation}) and, where necessary, sampling bias values from multiple training examples.

\subsubsection{Dilation} \label{subsubsec-sensitivity-dilation}

\begin{figure}
\centering
\includegraphics[width=\linewidth]{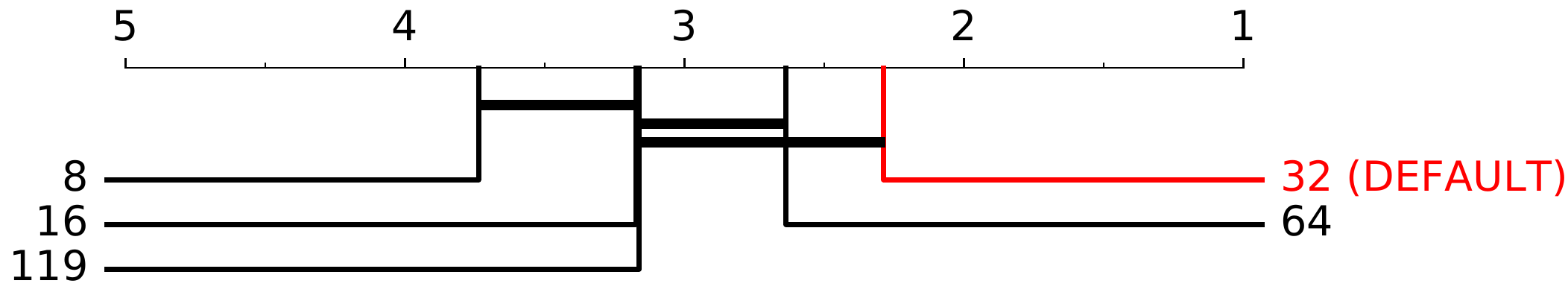}
\caption{Mean rank of different values for the maximum number of dilations per kernel.}
\Description[A maximum of 32 dilations per kernel, the default, ranks ahead of all other values]{A maximum of 32 dilations per kernel, the default, ranks ahead of all other values for the maximum number of dilations per kernel.  The pairwise differences are (with one exception) not statistically significant.}
\label{fig-sensitivity-dilation}
\end{figure}

Figure \ref{fig-sensitivity-dilation} shows the effect in terms of accuracy of different values for the maximum number of dilations per kernel.  The total number of features is kept constant, such that more features are computed per dilation for a smaller number of maximum dilations per kernel and vice versa: see Section \ref{subsubsec-dilation}.  There is little difference in accuracy between values of 16 and 119 (119 being the largest possible number of dilations per kernel for the default number of features, i.e., $\lfloor 10{,}000 / {84} \rfloor = 119$).  A value of 32 balances accuracy with the computational advantage of limiting the number of dilations per kernel, as discussed in Section \ref{subsubsec-dilation}.

\section{Conclusion}

We reformulate {\rocket} into a new method, {\minirocket}, making it up to $75$ times faster on larger datasets.  {\minirocket} shows that it is possible to achieve essentially the same accuracy as {\rocket} using a mostly-deterministic and much faster procedure.

{\minirocket} represents a significant advance in accuracy relative to computational cost.  {\minirocket} is much faster than any other method of comparable accuracy (including {\rocket}), and far more accurate than any other method of even roughly-similar computational expense.  Accordingly, we suggest that {\minirocket} should be considered and used as the default variant of {\rocket}.  We provide a na{\"i}ve facility for applying {\minirocket} to multivariate time series (available through the accompanying repository).  In future work, we propose to investigate more sophisticated approaches to multivariate time series, to explore the integration of {\minirocket} with nonlinear classifiers, and the use of {\minirocket} beyond time series data.

\begin{acks}
  This material is based on work supported by an Australian Government Research Training Program Scholarship, and the Australian Research Council under award DP190100017.  The authors would like to thank Professor Eamonn Keogh and all the people who have contributed to the UCR time series classification archive.  Figures showing mean ranks were produced using code from \citep{ismailfawaz_etal_2019}.
\end{acks}

\bibliographystyle{ACM-Reference-Format}
\bibliography{references}

\clearpage

\appendix

\section{Logistic Regression Training} \label{sec-appendix-training-details}

For each dataset, we shuffle the training set once, and train on increasingly large subsets of the shuffled training data.  The transform is performed in batches (of $2^{12}$ training examples), which are further divided into minibatches for training.  We use the same hyperparameters for all datasets: a validation set of $2{,}048$ examples, a minibatch size of 256, an initial learning rate $10^{-4}$, the learning rate is halved if validation loss does not improve after 50 updates, and training is stopped if validation loss does not improve after 100 updates.  The same hyperparameters can be reused for any dataset, because the classifier `always sees the same thing', i.e., blocks of transformed features.  Additionally, we cache the transformed features, in order to avoid unnecessarily repeating the transform when training for multiple epochs.

\section{Algorithm} \label{sec-appendix-pseudocode}

\begin{function}[h]
  \DontPrintSemicolon
  \SetCommentSty{textsf}
  \SetKwInOut{Input}{input}
  \SetKwInOut{Output}{output}

  \Input{\null\quad \makebox[1.25em]{$\boldsymbol{X}$\hfill:} time series \newline \null\quad \makebox[1.25em]{$D$\hfill:} dilations \newline \null\quad \makebox[1.25em]{$N$\hfill:} num features per dilation \newline \null\quad \makebox[1.25em]{$B$\hfill:} biases}

  \BlankLine

  \Output{\null\quad \makebox[1em]{$\boldsymbol{F}$\hfill:} features}

  \BlankLine

  \Begin{
  \tcp{indices of $\beta$ weights in kernels}
  $\boldsymbol{I} \leftarrow [[0,1,2], [0,1,3], \ldots, [6,7,8]]$ \;

  \BlankLine

  \For{$i \in [0, 1, \ldots, \textup{size}(\boldsymbol{X}) - 1]$}
  {

    \BlankLine

    $a \leftarrow 0$ \;

    \BlankLine

    $X \leftarrow \boldsymbol{X}[i]$ \;
    $A, G \leftarrow -X, 3X$ \tcp*{\S \ref{subsubsec-factoring-out}}

    \BlankLine

    \For{$j \in [0, 1, \ldots, | D | -1]$}
    {

      \BlankLine

      $\hat{p}_{0} \leftarrow j \ \text{mod} \ 2$ \;
      $p \leftarrow 4 \cdot D[j]$ \;

      \BlankLine

      precompute $C_{\alpha}, \hat{\boldsymbol{C}}_{\gamma}$ for $D[j]$ per \S \ref{subsubsec-all-kernels-at-once} \;

      \BlankLine

      \For{$k \in [0, 1, \ldots, 83]$}
      {

        \BlankLine

        $b \leftarrow a + N[j]$ \;

        \BlankLine

        $C_{\gamma} \leftarrow \hat{\boldsymbol{C}}_{\gamma}[\boldsymbol{I}[k]]$ \tcp*{\S \ref{subsubsec-all-kernels-at-once}}
        $C \leftarrow C_{\alpha} + C_{\gamma}$ \;

        \BlankLine

        $\hat{p}_{1} \leftarrow \hat{p}_{0} + k \ \text{mod} \ 2$ \tcp*{\S \ref{subsubsec-padding}}
        \lIf{$\hat{p_{1}}$}{$C \leftarrow C[p{:}| C | - p]$}

        \BlankLine

        $\boldsymbol{F}[i, a{:}b] \leftarrow \text{PPV}(C, B[a{:}b])$ \;

        \BlankLine

        $a \leftarrow b$ \;

        \BlankLine

      }
    }
  }
  \KwRet{$\boldsymbol{F}$}
  }
  \caption{transform($\boldsymbol{X},D,N,B$)}
  \label{pseudo-transform}
\end{function}

\begin{function}[h]
  \DontPrintSemicolon
  \SetCommentSty{textsf}
  \SetKwInOut{Input}{input}
  \SetKwInOut{Output}{output}

  \Input{\null\quad \makebox[1.25em]{$\boldsymbol{X}$\hfill:} time series \newline \null\quad \makebox[1.25em]{$n$\hfill:} num features \newline \null\quad \makebox[1.25em]{$m$\hfill:} max dilations per kernel}

  \BlankLine

  \Output{\null\quad \makebox[1.25em]{$D$\hfill:} dilations \newline \null\quad \makebox[1.25em]{$N$\hfill:} num features per dilation \newline \null\quad \makebox[1.25em]{$B$\hfill:} biases}
  \Begin{
  $\text{max} \leftarrow \log_{2} (\text{length}(\boldsymbol{X}) - 1) / 8$ \tcp*{\S \ref{subsubsec-dilation}}
  $\hat{D} \leftarrow [ \lfloor 2^{0} \rfloor, \lfloor 2^{\text{max}/m} \rfloor, \lfloor 2^{2 \cdot \text{max}/m} \rfloor, \ldots, \lfloor 2^{m \cdot \text{max}/m} \rfloor ]$ \;

  \BlankLine

  $D \leftarrow$ unique elements in $\hat{D}$ \;
  $\hat{N} \leftarrow$ count of each unique element in $\hat{D}$

  \BlankLine

  $N \leftarrow \lfloor \hat{N} \cdot n / 84 / m \rfloor$ \tcp*{scale to $n$, s.t. $\text{sum}(N) \approx n$}

  \BlankLine

  $r \leftarrow n - \text{sum}(N)$ \tcp*{apportion remainder}
  \lIf{$r > 0$}{$N[{:}r] \leftarrow N[{:}r] + 1$}

  \BlankLine

  $Q \leftarrow i \cdot \frac{1 + \sqrt{5}}{2} \ \text{mod} \ 1, \forall i \in [1, 2, \ldots, n]$ \tcp*{\S \ref{subsubsec-bias}}
  $B \leftarrow \text{sample}(\boldsymbol{X}, D, N, Q)$ \;

  \BlankLine

  \KwRet{$D, N, B$}
  }
  \caption{fit($\boldsymbol{X},n,m$)}
  \label{pseudo-fit}
\end{function}

\begin{function}[h]
  \DontPrintSemicolon
  \SetCommentSty{textsf}
  \SetKwInOut{Input}{input}
  \SetKwInOut{Output}{output}
  \Input{\null\quad \makebox[1.25em]{$\boldsymbol{X}$\hfill:} time series \newline \null\quad \makebox[1.25em]{$D$\hfill:} dilations \newline \null\quad \makebox[1.25em]{$N$\hfill:} num features per dilation \newline \null\quad \makebox[1.25em]{$Q$\hfill:} quantiles}

  \BlankLine

  \Output{\null\quad \makebox[1.25em]{$B$\hfill:} biases}

  \BlankLine

  \Begin{
  \tcp{indices of $\beta$ weights in kernels}
  $\boldsymbol{I} \leftarrow [[0,1,2], [0,1,3], \ldots, [6,7,8]]$ \;

  \BlankLine

  $a \leftarrow 0$ \;

  \BlankLine

  \For{$j \in [0, 1, \ldots, | D | - 1]$}
  {

    \BlankLine

    \For{$k \in [0, 1, \ldots, 83]$}
    {

      \BlankLine

      $b \leftarrow a + N[j]$ \;

      \BlankLine

      $X \leftarrow \text{random}(\boldsymbol{X})$ \;
      $A, G \leftarrow -X, 3X$ \tcp*{\S \ref{subsubsec-factoring-out}}

      \BlankLine

      compute $C_{\alpha}, \hat{\boldsymbol{C}}_{\gamma}$ for $D[j]$ per \S \ref{subsubsec-all-kernels-at-once} \;
      $C_{\gamma} \leftarrow \hat{\boldsymbol{C}}_{\gamma}[\boldsymbol{I}[k]]$ \;
      $C \leftarrow C_{\alpha} + C_{\gamma}$\;

      \BlankLine

      $B[a{:}b] \leftarrow \text{quantiles}(C, Q[a{:}b])$ \tcp*{\S \ref{subsubsec-bias}}

      \BlankLine

      $a \leftarrow b$ \;

      \BlankLine

    }
  }
  \KwRet{$B$}
  }
  \caption{sample($\boldsymbol{X},D,N,Q$)}
\end{function}

\end{document}